\documentclass[11pt]{article}
\usepackage[margin=1in]{geometry}
\usepackage[T1]{fontenc}
\usepackage{lmodern,microtype}
\usepackage{amsmath,booktabs,graphicx,natbib,xcolor,enumitem}
\usepackage[hidelinks]{hyperref}
\hypersetup{pdftitle={One Run Is Not Enough: Reminding a small Gemma 4 coding agent of its last finding made no measurable difference in a paired test.}}
\graphicspath{{../figures/}{figures/}}
\setlist{nosep,leftmargin=1.4em}
\newcommand{\code}[1]{\texttt{#1}}

\title{One Run Is Not Enough\\[0.35em]
{\large Reminding a small Gemma~4 coding agent of its last finding made no measurable difference in a paired test.}}
\author{Patrick Passos\\Renor}
\date{November 2026}

\begin{document}
\maketitle

\begin{abstract}
When a coding agent runs out of room, its earlier findings are cut and it forgets them. We tested a simple fix: automatically putting the agent's last finding back in front of it. We fixed the rules in advance and ran 98 coding tasks with and without this reminder, using a small Gemma~4 model. The reminder made no measurable difference: across all 98 tasks, 47 were solved with it and 48 without. Running the same task twice changed the outcome about one time in five, so a single run on a few dozen tasks cannot show whether memory helps.

In the treatment, the harness re-injects the explorer sub-agent's latest finding into every coder request, plus any notes the agent chose to write (it
almost never did). It does \emph{not} test agent-written memory.
We compare it with no memory using Gemma~4 31B (QAT,
4-bit) in the Gemma~4 Developer Agent harness~\citep{kaggle2026gemma4agent}, on 98 paired tasks
across three backends (Modal H100, 16 tasks; Kaggle 4$\times$L4, 32 tasks; AWS L40S, 50 tasks added in
a pre-registered extension). The primary outcome (a documented deviation from our preregistration), pooled run-1 resolve rate on the
first 48 pairs, was 29/48 with memory against 25/48 without: $+8.3$ points, Newcombe 95\% interval
$[-2.1, +18.3]$, exact McNemar $p=0.22$. The extension, decided after that result, pointed the other
way: 18/50 against 23/50, $-10.0$ points $[-20.2, +0.7]$, $p=0.125$. Combined (secondary, no
confirmatory claim), 98 pairs: 47/98 against 48/98, $-1.0$ points $[-8.3, +6.3]$, $p=1.0$. This is a
null result, and the sign did not replicate. With 19\% of tasks flipping between two identical
no-memory runs, the combined 95\% interval excludes harms larger than about 8 points and gains larger than about 6, and the design had 80\% power only for effects of about $\pm12$ points or more. The registered secondary, repeated actions
per compaction, was also null ($+0.30$ $[-4.50, +5.98]$). Two directional hypotheses found on Modal
did not replicate on Kaggle. Pre-registered replicate runs of every task (Phase~R) are reported separately. The mechanism findings (first 48 tasks) are clearer. The model almost never wrote notes
(2 of 16 and 4 of 32 runs). Memory costs context: the memory arm's first request is larger, and it
compacted more on Modal. Something was injected at only 47\% to 65\% of compactions. On two Modal
runs, memory coincided with a \code{submit\_patch} loop. We also report an A/A calibration and a
validity funnel (129 tasks, of which 105 were valid). Together they show that single-run memory
claims on task sets of this size sit inside run-to-run noise.
\end{abstract}

\section{Introduction}
The promise of agent memory is that context told once stays available to an agent when it is relevant.
For a small coding agent, the first place that promise breaks is inside a single task. The Gemma~4
harness compacts the conversation when the previous prompt reaches 14,336 tokens. In 16 no-memory
throughput traces, 13 tasks compacted, and 53 of 83 exact repeats of a read, a command or an
explorer question came after a compaction.

Experience memory across tasks~\citep{shinn2023reflexion,zhao2023expel,ouyang2025reasoningbank}
targets a different problem. For coding agents the evidence is sobering:
VibeMemBench~\citep{fan2026vibemembench} found that 11 of 12 solver and memory pairings did not
beat memory-off. We therefore built the study expecting a likely null result and designed it to
report one honestly.

\paragraph{Question.} Does automatically re-injecting the explorer's last finding after compaction
help Gemma~4 31B resolve real repository tasks?

\paragraph{Contributions.}
\begin{enumerate}
\item A paired comparison, stratified by backend, with a null primary result, a
  pre-registered extension whose sign differs, and intervals that state which effects they exclude
  (Section~\ref{sec:primary}).
\item Mechanism findings: adherence (the model rarely writes memory), the context cost of memory,
  delivery rates, and a treatment-associated \code{submit\_patch} loop (Section~\ref{sec:mech}).
\item A measurement method for small agent studies: A/A calibration, a validity funnel, smoke
  gates, frozen detectors, byte-identical baseline builds, and a timestamped change log
  (Section~\ref{sec:trust}).
\end{enumerate}

\section{Method}
\subsection{Setting}
The model is Gemma~4 31B (QAT, w4a16, thinking budget 4096) in the competition's declarative ADK
harness: a coder agent, an \code{explorer} sub-agent, a 100-tool-call cap and a 60-minute cap. Tasks
are the 129 public development tasks (fastapi 67, rich 48, requests 13, httpx 1), in the
SWE-bench~\citep{jimenez2023swebench} family. Scoring uses the harness's own \code{verify\_task} in
a per-task Python 3.13 environment built by the harness's Docker setup steps. The agents run under
the subprocess sandbox, not the official Docker scorer.

\subsection{Arms}
\begin{itemize}
\item \textbf{none}: the baseline submission. Its build is byte-identical across every code pin
  (a test rebuilds it at each pin and compares the files).
\item \textbf{recap}: after every compaction, the coder's instruction carries two session-state slots
  that ADK fills on each model call: \code{\{explorer\_finding?\}}, the explorer's latest answer (at
  most 15 lines), and \code{\{working\_notes?\}}, a page of notes the agent wrote through a
  \code{notes} sub-agent, capped at 1,500 tokens. The agent is asked to write a note after each
  explorer answer and each edit.
\end{itemize}
The arm began as agent-requested recall. In three smoke checks (6 runs), no run wrote a note, so
before any result on the 48 tasks we moved recall to automatic injection and then added the
explorer finding. This was the pre-registered fallback for near-zero note writes. \textbf{What we
test is therefore automatic carry-over of the explorer's last finding, plus rare notes. We do not
test agent-written working memory.} No embeddings are used.

\subsection{Pre-registration and analysis}
A hash-frozen pre-registration and a timestamped protocol change log record each decision with its
time. The run-1 pooled form of the primary was fixed 5 minutes before the first results commit. The headline moved from cross-task lessons to the in-task recap. That
decision was made from no-memory traces only, before any memory-arm run.
\begin{itemize}
\item \textbf{Primary:} resolve rate, intention to treat, run 1 against run 1, pooled over 48 pairs.
  We report a backend-stratified bootstrap (10,000 resamples), the Newcombe paired score interval
  and exact McNemar. Run 2 and per-task means are sensitivity analyses that cannot change the
  headline.
\item \textbf{Key secondary:} repeated actions per compaction. A repeat is a root-agent tool call,
  after the first compaction, whose tool and normalized arguments match an earlier call. We report
  the paired mean difference with a bootstrap CI and a Wilcoxon signed-rank test.
\item \textbf{Replication:} two directional hypotheses (Section~\ref{sec:repl}) were discovered on
  Modal and frozen as code. They were then tested on Kaggle before any Kaggle recap row was read.
\item \textbf{Adherence and delivery:} notes written, and the share of compactions at which something
  was injected. Both were registered as outcomes.
\item \textbf{Extension:} after seeing the 48-pair result, we pre-registered 50 more tasks on a third
  backend (commit bdf6c0e3, before any extension row): one run per arm at the frozen pins, with its
  own two-sided primary (exact McNemar, Newcombe interval). The combined 98-pair analysis,
  stratified by backend, is secondary. Pre-registered replicate runs of every task (Phase~R) are reported separately.
\end{itemize}
Every test not listed above is post-hoc and labelled so. We ran 23 post-hoc tests in all
(Appendix~\ref{app:posthoc}).

\subsection{Making the measurement trustworthy}\label{sec:trust}
\paragraph{Validity funnel.} In the public environment, only 37 of 129 tasks passed with their gold
patch (the reference fix). Adding the missing test dependencies raised this to 68. Installing, for
each snapshot, the newest competition wheel its own pyproject allows raised it to 105. That step
targeted the 44 fastapi gold failures caused by a too-new starlette. The 24 exclusions follow
written rules. The 48-task subset was drawn once, with a seed (20261107), from the 105 valid tasks,
stratified by repository. The 50 extension tasks are the remaining valid tasks, minus 7 used for
tuning, smoke and throughput checks.

\paragraph{Smoke gates.} Every arm had to pass small out-of-sample checks before running on the 48
tasks. These checks caught three bugs that would otherwise have produced results that looked
plausible but were wrong:
(1)~\emph{plumbing}: the first skill check was rejected, because its write was flagged as path
traversal and the skill was neither declared nor reachable;
(2)~\emph{sandbox paths}: under the subprocess sandbox, skill arguments named the host's
\code{/workspace} and \code{/tmp}, so memory reported itself unavailable, and notes would have been
shared across concurrent tasks;
(3)~\emph{adherence}: agents skipped recall and wrote no notes, so the arm would have run as
``memory'' with nothing in it.
A CPU replay check also caught an interpreter path that would have failed every task in a fresh
session. It showed that, after masking random sandbox ids, the no-memory arm sends 33 of 33 model
requests identical before and after the memory plumbing was added.

\paragraph{A/A calibration.} We compared the no-memory arm against itself, run 2 against run 1
(Table~\ref{tab:aa}). Of the 48 tasks, 9 changed outcome (18.75\%), with no treatment at all. With
that discordance, the paired standard error over 48 pairs is about 0.063, so 80\% power at
$\alpha=0.05$ requires effects of about $\pm18$ points.

\begin{table}[t]\centering\small
\caption{Noise floor with no treatment (no memory, run 2 vs run 1).}\label{tab:aa}
\resizebox{\linewidth}{!}{%
\begin{tabular}{lll}\toprule
 & Kaggle 32 & Modal 16 \\\midrule
resolved, run 1 / run 2 & 14 / 16 & 11 / 12 \\
difference [95\% CI] & $+0.063$ $[-0.094, +0.219]$ & $+0.063$ $[-0.125, +0.250]$ \\
McNemar & $b=4$, $c=2$, $p=0.69$ & $b=2$, $c=1$, $p=1.0$ \\
repeat-rate difference [95\% CI], $n$ & $-1.97$ $[-7.38, +2.70]$, 26 & $+0.27$ $[-1.24, +1.58]$, 14 \\
leaked thought text, runs (run 1 / run 2) & 5 / 3 & 0 / 0 \\
overflows (run 1 / run 2) & 5 / 4 & 1 / 0 \\\bottomrule
\end{tabular}}
\end{table}

\section{Results}
\subsection{Resolve rate: primary, extension and combined}\label{sec:primary}

\begin{table}[t]\centering\small
\caption{Resolve rate. Run 1, intention to treat. The backend-stratified bootstrap gives
$[0.0, +18.8]$ pooled for the 48, $[-20.0, 0.0]$ for the extension and $[-8.2, +6.1]$ combined. The
bootstrap bound at 0 follows from the 5/1 and 1/6 splits and is not evidence for or against harm.}\label{tab:primary}
\begin{tabular}{lccccc}\toprule
Set & recap / none & discordant & difference & Newcombe 95\% & McNemar $p$ \\\midrule
Pooled, 48 (primary) & 29 / 25 & 5 / 1 & $+8.3$ pts & $[-2.1, +18.3]$ & 0.22 \\
Modal H100, 16 & 12 / 11 & 2 / 1 & $+6.3$ pts & $[-17.5, +29.3]$ & 1.0 \\
Kaggle 4$\times$L4, 32 & 17 / 14 & 3 / 0 & $+9.4$ pts & $[-2.1, +20.2]$ & 0.25 \\
AWS L40S, 50 (extension primary) & 18 / 23 & 1 / 6 & $-10.0$ pts & $[-20.2, +0.7]$ & 0.125 \\
Combined, 98 (secondary) & 47 / 48 & 6 / 7 & $-1.0$ pts & $[-8.3, +6.3]$ & 1.0 \\\bottomrule
\end{tabular}
\end{table}

\begin{figure}[t]\centering
\includegraphics[width=0.92\linewidth]{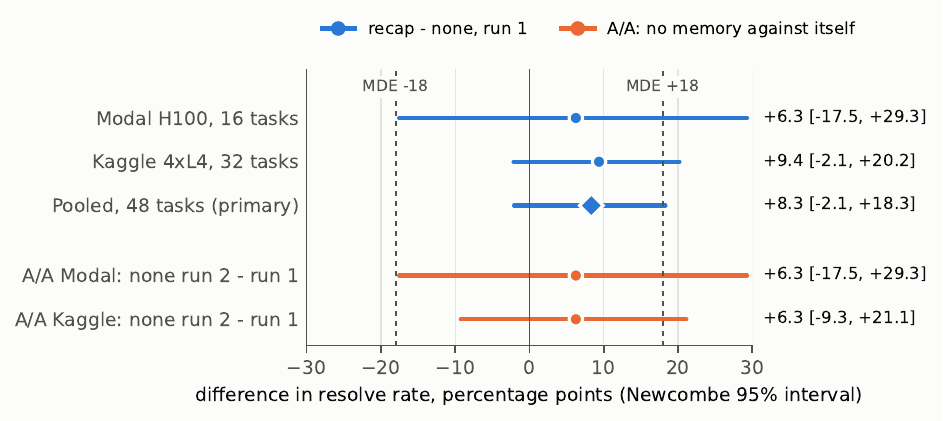}
\caption{Resolve-rate difference with Newcombe 95\% intervals. Blue: recap minus none, per backend
and pooled. Orange: the no-memory arm against itself. Dashed: the minimal detectable effect.}
\label{fig:forest}
\end{figure}

On the first 48 tasks, memory resolved 4 more (Table~\ref{tab:primary}). This is not a significant
effect.
Figure~\ref{fig:forest} puts this next to the A/A contrast. The no-memory arm, compared with
itself, moved by the same $+6.3$ points as the Modal treatment effect. Per-protocol, which excludes
runs flagged \code{plumbing\_failed} (11 per backend), the result is the same (26 pairs, 13 vs 11,
$p=0.5$). So is the registered sensitivity analysis, the per-task mean over all available runs
($+5.2$ points, bootstrap $[-3.1, +12.5]$).

The six discordant pairs of the 48 look like noise (post-hoc, descriptive). The no-memory arm solved all 5 of
its run-1 losses in its own run 2. The one task it won, requests\_7328, was a recap run caught in
the \code{submit\_patch} loop (Section~\ref{sec:mech}).

\paragraph{Extension.} On 50 new tasks the estimate reversed: memory resolved 18 and no memory 23,
with 6 tasks solved only without memory and 1 only with it. The interval includes 0 and the exact
McNemar $p$ is 0.125, so this is not significant by the registered test either. (Descriptive, not registered: in the extension, run 1 hit a context overflow in 12 of 50 no-memory runs and 14 of 50 memory runs, and 13 memory runs were flagged \texttt{plumbing\_failed}. We make no claim about the cause.) Neither estimate
exceeds the minimal detectable effect (about 18 points for 48 pairs, 17 for 50), and we registered
no test of heterogeneity between sets, so the first result is not a stable finding. The combined
estimate is $-1.0$ point $[-8.3, +6.3]$. It is secondary, pools backends and makes no confirmatory
claim. We read it as no evidence of an effect in either direction. Its 95\% interval excludes harms larger than about 8 points and gains larger than about 6; the design had 80\% power only for effects of about 12 points or more.

\subsection{Key secondary: repeated actions per compaction (registered)}

\begin{table}[t]\centering\small
\caption{Repeats per compaction, recap $-$ none: paired mean difference [bootstrap 95\% CI],
Wilcoxon $p$, and tasks with a rate in both arms.}\label{tab:rep}
\begin{tabular}{llll}\toprule
 & stratum & difference [95\% CI] & Wilcoxon $p$, $n$ \\\midrule
registered & pooled & $+0.30$ $[-4.50, +5.98]$ & 0.27, 41 \\
 & Modal & $+8.54$ $[-0.74, +22.48]$ & 0.28, 15 \\
 & Kaggle & $-4.45$ $[-9.34, -1.22]$ & 0.019, 26 \\\midrule
post-hoc: post-submit & pooled & $-1.35$ $[-2.46, -0.33]$ & 0.23, 41 \\
\quad submits removed & Modal & $+0.21$ $[-1.15, +1.39]$ & 0.25, 15 \\
 & Kaggle & $-2.26$ $[-3.81, -0.84]$ & 0.029, 26 \\\bottomrule
\end{tabular}
\end{table}

The registered pooled statistic is null, and its 10\% trimmed mean is $-1.17$ (Table~\ref{tab:rep}).
The per-backend values are descriptive, since the registration is pooled, and each is driven by
\code{submit\_patch} loops (Section~\ref{sec:mech}). Modal's $+8.54$ comes from two looping recap
runs. Kaggle's $-4.45$ includes one looping no-memory run, and it shrinks to $-2.26$ once
post-submit submits are removed.

\paragraph{Post-hoc denominator checks (exploratory).} A per-compaction rate can fall simply because
compactions rise. We therefore also looked at raw repeats per task, repeats per tool call after the
first compaction, and compactions per task. These checks belong to the 23 post-hoc tests
(Appendix~\ref{app:posthoc}). Only single-stratum tests fell below 0.05: the Kaggle repeat rate with
post-submit submits removed ($p$ 0.029), Kaggle repeats per call (0.021 registered steps, 0.042
without post-submit submits), Modal compactions per task (0.032), and the Modal discovery of H-b
(0.0088). None survives a Bonferroni correction over the 23, and no pooled test reached 0.05.
Pooled raw repeats per task did not fall (both post-hoc): $+8.1$ $[-18.0, +35.9]$, $p$ 0.54 at the registered steps, inflated by the loops, and $-0.98$ $[-3.17, +1.02]$, $p$ 0.53 with post-submit submits removed. The pooled post-hoc
rate has a bootstrap interval that excludes 0, but its rank test does not reach 0.05 ($p$ 0.23).
We do not call any of these results significant. A ``memory reduces repeated actions'' claim is not
supported.

\subsection{Replication of two Modal hypotheses (pre-registered)}\label{sec:repl}
\textbf{H-a:} recap has more compactions before the first successful submit. \textbf{H-b:} recap has
more runs that leak raw \code{<|channel>} thought text into the coder's output. Both were discovered
on Modal and frozen as code before any Kaggle recap result was read.

\begin{table}[h]\centering\small
\caption{Neither hypothesis replicated.}\label{tab:repl}
\begin{tabular}{lll}\toprule
 & Modal (discovery, post-hoc there) & Kaggle (replication) \\\midrule
H-a, tasks recap more / fewer / equal & 8 / 2 / 6, one-sided $p$ 0.059 & 17 / 11 / 4, $p$ 0.16 \\
H-b, runs with leaked thought, recap / none & 6 / 0 of 16, Fisher $p$ 0.0088 & 5 / 5 of 32, $p$ 0.63 \\\bottomrule
\end{tabular}
\end{table}

The H-b base rate differs by backend, and we recorded this before the replication. Kaggle's
no-memory arm already leaked in 5 and 3 of 32 runs, against 0 of 16 on Modal. A different serving
stack is the likeliest explanation for Modal's 6/0. Memory is not.

\subsection{Mechanism (first 48 tasks)}\label{sec:mech}

\begin{figure}[t]\centering
\includegraphics[width=0.92\linewidth]{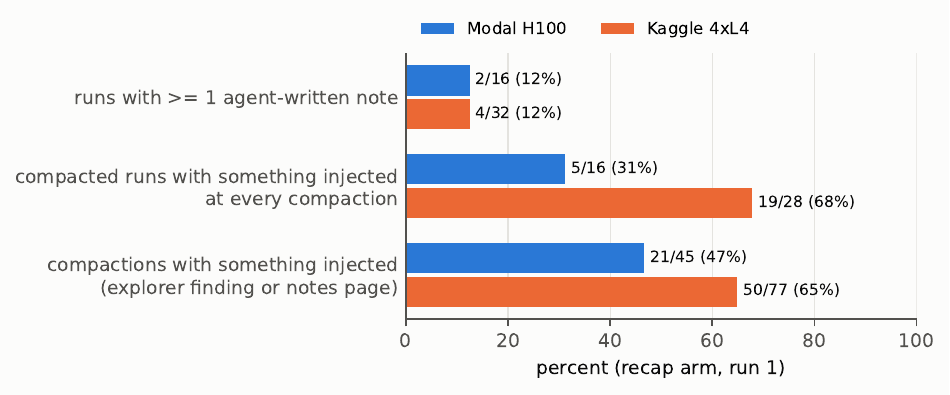}
\caption{Adherence and delivery, recap arm, run 1.}\label{fig:adh}
\end{figure}

\paragraph{Adherence.} The model almost never writes memory (Figure~\ref{fig:adh}). Two of 16 Modal
runs wrote any note (2 notes in all), and 4 of 32 Kaggle runs did (5 notes). In the six smoke runs
it wrote none. All 11 \code{plumbing\_failed} runs on Modal were adherence failures, not machinery
failures: 4 never ran the first recall, and 7 had nothing to inject at a compaction. Renor's product
telemetry, which is internal and not independently published, shows the same pattern: agents ask
for memory on about 1 in 24 prompts unless recall is automatic. Memory that depends on the agent's
cooperation largely does not happen.

\paragraph{Delivery.} Even automatic injection needs the agent to have produced something. On Modal,
something was injected at 21 of 45 compactions (47\%); on Kaggle, at 50 of 77 (65\%). The mean share
per compacted run was 38\% and 70\%. Before a run's first explorer call, nothing can be injected.

\paragraph{The context cost of memory.} The memory section, the skill and the notes tool add about
2,000 tokens to every request. The median first coder request was 3,966 tokens against 2,311
(Modal). The recap arm therefore reaches the compaction trigger sooner and restarts larger. The
trigger reads the previous call's prompt size, so one large step can push the input past 16,384
tokens, where 16,384 output tokens meet the 32,768-token limit, before compaction runs.
Table~\ref{tab:desc} gives the per-arm counts. Compactions rose from 23 to 45 on Modal (post-hoc,
per task $+1.38$ $[+0.31, +2.69]$) and from 71 to 77 on Kaggle. Kaggle overflows were 7 against 5
(3 against 2 before any compaction); Modal overflows were 0 against 1.

\begin{table}[t]\centering\small
\caption{Descriptive counts, run 1, per backend and arm.}\label{tab:desc}
\begin{tabular}{lcccc}\toprule
 & Modal none & Modal recap & Kaggle none & Kaggle recap \\\midrule
runs & 16 & 16 & 32 & 32 \\
resolved & 11 & 12 & 14 & 17 \\
compactions & 23 & 45 & 71 & 77 \\
context overflows (before first compaction) & 1 (1) & 0 (0) & 5 (2) & 7 (3) \\
tool-call cap hits & 1 & 1 & 2 & 0 \\
\code{submit\_patch} loops & 0 & 2 & 1 & 0 \\
notes written (runs with a note) & -- & 2 (2) & -- & 5 (4) \\\bottomrule
\end{tabular}
\end{table}

\paragraph{A treatment harm: the \code{submit\_patch} loop.} In 2 of 16 Modal recap runs,
\code{submit\_patch} returned ok, but the agent then called it 441 and 385 more times with the same
patch, until the 500-turn limit. The harness ends a run only on a response with visible text and no
tool call. No Modal no-memory run passed 120 model calls. The loops do not change the resolve
outcome, but they waste the budget and account for 879 of the recap arm's run-1 Modal repeats.
Kaggle had one loop, and it was in the no-memory arm, so the loop is not specific to memory.

\section{Related work}
\textbf{Experience memory} stores lessons across tasks: Reflexion~\citep{shinn2023reflexion},
ExpeL~\citep{zhao2023expel}, Agent Workflow Memory~\citep{wang2024awm},
ReasoningBank~\citep{ouyang2025reasoningbank}, Dynamic Cheatsheet~\citep{suzgun2025dynamiccheatsheet}
and, for coding, SWE-Exp~\citep{chen2025sweexp}, subtask-level memory~\citep{shen2026subtask} and
Agent KB~\citep{tang2025agentkb}. VibeMemBench~\citep{fan2026vibemembench} evaluates memory systems
for coding agents on real repositories, and 11 of 12 pairings did not beat memory-off; our result
fits that pattern, for in-task memory and a smaller model. \textbf{Context management} treats the
window as a resource: MemGPT~\citep{packer2023memgpt} pages memory like virtual memory, and
SWE-MeM~\citep{gao2026swemem} learns in-episode compression with reinforcement learning. Our arm is
untrained and needs no cooperation from the agent for the explorer channel. Conversational memory
systems~\citep{chhikara2025mem0,rasmussen2025zep,latimer2025hindsight} and their
benchmarks~\citep{wu2024longmemeval} target dialogue rather than code. \textbf{Structure without
memory}, such as Code Isn't Memory~\citep{bhola2026codeisntmemory},
RepoGraph~\citep{ouyang2024repograph}, CodexGraph~\citep{liu2024codexgraph} and
LocAgent~\citep{chen2025locagent}, is already partly in the harness through its code-graph tool.
Agent scaffolds~\citep{yang2024sweagent,xia2024agentless} are the kind of base agent memory sits on.
We cite no benchmark number as comparable to this harness.

\section{Threats to validity}
\begin{itemize}
\item \textbf{Backend is confounded with repository.} Seeded round-robin plus every-third-task
  allocation sent all 9 requests tasks to Modal (Modal: requests 9, rich 4, fastapi 3; Kaggle:
  fastapi 16, rich 15, httpx 1). Pairing within a task keeps each difference valid, but no
  per-backend or per-repository effect can be claimed.
\item \textbf{Run noise, one model.} Only run 1 is primary. The extension has one run per arm, and
  recap run 2 exists only on Kaggle. Pre-registered replicate runs of every task (Phase~R) are reported separately. The extension was decided
  after the 48-pair result and runs on a different backend (fastapi 26, rich 24 tasks). All results
  are for Gemma~4 31B QAT.
\item \textbf{Valid-set skew.} 105 of 129 tasks were valid. The excluded tasks are mostly fastapi
  tasks that error on their base commit.
\item \textbf{Sandbox.} Agents ran under the subprocess sandbox with our scorer-like environment, not
  the official Docker scorer.
\item \textbf{Not the Renor product.} Renor's retrieval, graph and cross-session memory were not
  tested.
\item \textbf{No embeddings}, no misleading-memory control (it was dropped under the \$25 Modal
  cap), and a modest ceiling: no memory resolved 23 of 32 Modal runs.
\end{itemize}

\section{Conclusion and future work}
Automatically carrying the explorer's last finding across compaction did not measurably change how
often Gemma~4 31B resolved tasks. The first 48 pairs favored memory by 8.3 points, 50 new pairs
favored no memory by 10.0, and the combined 98 differ by $-1.0$. The combined 95\% interval excludes harms larger than about 8 points and gains larger than about 6, and the design had 80\% power only for effects of about 12 points or more.
Its clearest lessons are about the mechanism and the measurement. Small models do not write memory
unless memory writes itself. Memory costs context, which in turn causes more compactions. With
19\% of tasks flipping between identical runs, a single-run memory comparison on a few dozen tasks
cannot tell a real effect from noise.

Two changes come next. The first is a recap that the harness builds from the tool history, so it
needs no cooperation from the agent and is present at every compaction. The second, for the main
competition, is a guard that ends a run after the first successful \code{submit\_patch}, in every
arm.

\paragraph{Reproducibility.} \code{results/run\_rows.jsonl} holds one row per run, with outcomes
and measures only. \code{python results/analyze\_rows.py} regenerates every number. The notebook
redraws both figures. \code{paper/check\_numbers.py} checks each number in the Writeup against the
analysis output, the rows and the change log. Code, data and the commit history of the pre-registration:
\url{https://github.com/renorai/gemma4-memory-study}.

\paragraph{AI assistance.} AI coding agents (Anthropic Claude and OpenAI Codex) helped the author implement and run the experiments, write the analysis code and draft this text. The author directed the work, reviewed the manuscript and takes full responsibility for its contents.

\bibliographystyle{plainnat}
\bibliography{refs}

\appendix
\section{All post-hoc tests}\label{app:posthoc}
All 23 are computed by \code{results/analyze\_rows.py}. The Modal worker's own H-a counter (9/2/5, two-sided $p$ 0.061) tested the same hypothesis on the same traces as test 22 (8/2/6 by the frozen detector) and is not counted separately. Also not counted: the primary, the
registered repeat rate, the Kaggle replications of H-a and H-b, and the A/A calibration. All
Wilcoxon tests are recap $-$ none, paired by task.

\begin{center}\small
\resizebox{\linewidth}{!}{%
\begin{tabular}{rllr}\toprule
\# & test & stratum & $p$ \\\midrule
1--3 & repeat rate, post-submit submits removed & pooled / Modal / Kaggle & 0.233 / 0.251 / 0.029 \\
4--6 & (a) raw repeats per task, registered steps & pooled / Modal / Kaggle & 0.542 / 0.236 / 0.108 \\
7--9 & (a) raw repeats per task, no post-submit submits & pooled / Modal / Kaggle & 0.526 / 0.157 / 0.109 \\
10--12 & (b) repeats per call after compaction, registered steps & pooled / Modal / Kaggle & 0.060 / 1.000 / 0.021 \\
13--15 & (b) the same, no post-submit submits & pooled / Modal / Kaggle & 0.135 / 0.922 / 0.042 \\
16--18 & (c) compactions per task & pooled / Modal / Kaggle & 0.090 / 0.032 / 0.548 \\
19--21 & per-protocol repeat rate, post-hoc variant & pooled / Modal / Kaggle & 0.712 / 0.625 / 0.391 \\
22 & H-a one-sided Wilcoxon (Modal discovery) & Modal & 0.059 \\
23 & H-b one-sided Fisher (Modal discovery) & Modal & 0.0088 \\\bottomrule
\end{tabular}}
\end{center}

\end{document}